# DICTIONARY-BASED CONCEPT MINING: AN APPLICATION FOR TURKISH

Cem Rıfkı Aydın[1], Ali Erkan[1], Tunga Güngör[1] and Hidayet Takçı[2]

[1]Department of Computer Engineering, Boğaziçi University, Istanbul, Turkey
cemrifkiaydin@gmail.com, alierkan@gmail.com, gungort@boun.edu.tr
[2]Department of Computer Engineering, Cumhuriyet University, Sivas, Turkey
htakci@gmail.com

## ABSTRACT

*In this study, a dictionary-based method is used to extract expressive concepts from documents. So far, there have been many studies concerning concept mining in English, but this area of study for Turkish, an agglutinative language, is still immature. We used dictionary instead of WordNet, a lexical database grouping words into synsets that is widely used for concept extraction. The dictionaries are rarely used in the domain of concept mining, but taking into account that dictionary entries have synonyms, hypernyms, hyponyms and other relationships in their meaning texts, the success rate has been high for determining concepts. This concept extraction method is implemented on documents, that are collected from different corpora.*

## KEYWORDS

*Concept mining, Morphological analysis, Morphological disambiguation*

## 1. INTRODUCTION

Concept can be thought of as a general idea or something specific conceived through the mind. A human can easily have a general opinion concerning something which he/she is reading, watching or to which she is listening. But as for computers, since they lack the functionality of human brain that can easily classify many patterns and have conceptual ideas on an object, they have to process many statistical or machine learning methods.

Concept mining is a field of study where text, visual or audio materials are processed and general concepts are extracted. Amongst its applications, the most commonly used approach for extracting concepts is concerned with text materials. In this paper we are interested in extracting concepts from textual documents. Concept mining is a hard and very beneficial operation. In many areas it can be used in an automatic, efficient and computerized manner. For example it can ease categorizing the judiciary classes, such as adult court, appellate court and many others [1]. Also in medical field, it can help both the patients and doctors in many different ways [2, 3]. Similarly, banks and other financial institutions may use concept mining to track the profiles of some creditworthy customers.

During last decades, through the enhancement in natural language processing, many artificial intelligence methods have been developed and used for extracting concepts, such as Latent Dirichlet Allocation (LDA) that aims to determine topics of a textual material. Besides those methods, the most widely used lexical database for this field of study is WordNet. In WordNet the relationships between the words are stored in groups called synsets. These sets include hypernymy, hyponymy, synonymy relationships which are the most important and commonly used ones amongst many others. Hypernymy of a word can give us a general meaning of that word, hereby this relational property is the most widely used one for concept mining. This database is widely used for English concept extraction, but as for languages that are not as widely spoken as English, WordNet is not properly built and many word relationships (such as hypernymy) lack. That is why in these languages, the use of this lexical database would not yield as successful and meaningful results as it does in English.

In this paper, we used a novel approach for extracting concepts in Turkish through the use of a dictionary. Dictionaries have word entries, meaning texts, lexical class (noun, adjective, etc.) and some other properties. The meaning text of a word does, in fact, include its many relationally-relevant words, such as hypernyms or synonyms. So it is useful to benefit from the meaning texts in this dictionary to extract concepts. But these dictionary data are in an unstructured form, so they are pre-processed in this work, then concept mining method is run. The dictionary that is used is TDK (Türk Dil Kurumu - Turkish Language Association) Dictionary, which is the official Turkish dictionary that is the most respectful, accurate and most widely used one. We used the *bag-of-words* model for this study and we developed an algorithm which takes the frequencies of words in the document into account, since concepts do generally have to do with the most frequent words and the words related to those words present in the textual material. The success rate we observed is high compared with the other works done in concept mining for Turkish.

The remainder of the paper is as follows. In Section 2 related work concerning this field of study is examined. Section 3 elaborately examines the method developed in this work. Section 4 gives the evaluation results and the success rates achieved with the proposed method. Finally, the conclusion and future work are given in Section 5.

## 2. RELATED WORK

Numerous comprehensive studies are done in widely used languages in the domain of concept mining. In studies carried out in this domain, generally machine learning algorithms are used. Also WordNet, a lexical database, is commonly used. As for Turkish, the concept mining domain is still immature and also in this language some machine learning methods are used to extract concepts.

In one study, it is proposed that using WordNet's synset relations and then implementing clustering process may help obtaining meaningful clusters to be used for concept extraction [4]. The synset relationships are used as features for clustering documents and then these are used for succeding clustering. But in this study word disambiguation has not been performed and it is observed that using synset words has decreased the clustering success.

There have been also developed some toolkits for concept mining, one of which is ConceptNet [5]. In accordance with this toolkit, similar to the WordNet synset relationships, spatial, physical, social, temporal and psychological aspects of everyday life are taken into account. In this toolkit, concepts are extracted by processing these relationships and a data structure is built. For instance a hierarchical relationship is named "IsA" and in a document the counterpart relational word of the word "apple" may be "fruit" in accordance with this relationship. If there are many common words that are extracted through relationships in a document, by taking also their frequencies into accounts, successful results may be achieved. ConceptNet is much richer than WordNet in terms of its relational structure.

As for Turkish, there have been a few studies concerning this study. It was proposed that meaningful concepts could be extracted without the use of a dictionary and with the clustering method processed on documents in corpora [6]. Here clusters would have initially been assigned concepts and the documents to which clusters are assigned, would have those concepts.

In an another study, a method is developed for extracting concepts from web pages [7]. The frequencies of words in the document are taken into account and words are assigned different scores according to their html tags. Words between specific tags such as "<b>" and "<head>" are assigned higher scores. Then the words exceeding specified thresholds are determined as concepts. Success rates are reported to be high.

A study is done concerning extracting concepts from constructing digital library, documents in this library are categorized through clustering in accordance with those concepts [8]. An equation is created that multiplies term frequency (tf), inverse document frequency (idf), diameter, length, position of the first occurrence, and distribution deviation values of the keywords. Whichever words give the highest scores in accordance with this equation, they are selected as probable concepts. A higher success rate is achieved as compared with methods that take only tf and idf into account used for extracting concepts.

There have been used mostly lexical, relational databases and clustering methods for Concept Mining, besides those also Latent Dirichlet Allocation [9] and some other artificial intelligence methods have been used for extracting topics and concepts. Our method uses a statistical method that makes use of a dictionary, which has not been used in Turkish so far.

## 3. METHODOLOGY

In the concept mining domain, several dictionaries and lexical databases such as WordNet are used. Structural and semantic relationships between words can give us general idea (concept) about the words. In WordNet, especially hypernymy relation [10] is preferred in concept extraction, since it is the most relevant relation to a generalization of a word. There are also other relationships in WordNet such as meronymy and synonymy, those relationships give us other semantic relevances that may be useful in determining the concept of a word. For example, the synonym of the word "attorney" is "lawyer" and if a document from which concepts are to be extracted has many occurrences of the word "attorney", one probable concept may be "lawyer".

So far, the use of WordNet in Concept Mining has been extremely dominant. Taking it into account, in this work we used a novel approach, that is we used dictionary to help extract concepts from documents.

### 3.1. Dictionary Structure

The dictionary we used in this work includes the words and the meaning text of those words, where the words in the meaning carry specific relationships with each other. Figure 1 shows the XML structure of a dictionary item.

```
<entry>
    <name> jaguar </name>
    <affix> undefined </affix>
    <lex_class> isim, zooloji </lex_class>
    <stress> undefined </stress>
    <pronunciation> Fransızca jaguar </pronunciation>
    <origin> Fransızca </origin>
  - <meaning>
        <meaning_class> undefined </meaning_class>
        <meaning_text>  Kedigillerden, Orta ve Güney
            Amerika'da yaşayan, postu iri benekli memeli
            türü (Felis onca).</meaning_text>
      - <quotation>
            <author> undefined </author>
            <quotation_text> undefined </quotation_text>
        </quotation>
    </meaning>
    <atasozu_deyim_bilesik> undefined </atasozu_deyim_bilesik>
    <birlesik_sozler> undefined </birlesik_sozler>
</entry>
```

Figure 1. Structure of a dictionary entry, "jaguar", in XML format

In Figure 1, the tag "<atasozu_deyim_bilesik>" stands for "proverb, idiom, compound" and the tag "<birlesik_sozler>" stands for "compound phrases" in Turkish. In this work we took only "name", "lex_class" and "meaning" tag elements into account. We processed only the words,

"lex_class" of which are nouns, also analogies are made based on the meaning text nouns. There may be several meanings of a word, we took all of them into account and selected only one of them after disambiguation. The main relationships that would be encountered in dictionary item meaning texts can be summarized as follows:

- Synonymy: It is a relation that two words have equivalent meanings. It is a symmetrical relationship. (For example the words "intelligent" and "smart" have this relationship.)
- Meronymy: It is a relation that one of the words is a constituent of the other word. It is not a symmetrical relationship. (For example the words "hand" and "finger have this relationship.)
- Location: It is a relation that shows the location of a word with respect to the other word. (For example the words "kitchen" and "house" have this relationship.)
- Usability: It is a relation that one word is used for an aim. (For example "toothbrush" is used for "brushing teeth".)
- Effect: It is a relation that one action leads to a result. (For example taking medication leads to a healthy state.)
- Hypernymy: It is a relation that one word is a general concept of an another word. (For example the word "animal" is the hypernym of the word "cat".)
- Hyponymy: It is a relation that one word is a more specific concept of an another word. (For example the word "school" is hyponym of the word "building".)
- Subevent: It is a relation that one action has a sub-action. (For example waking up in the morning would make one yawn.)
- Prerequisite relation: It is a relation that one action is a prerequisite condition for another one. (For example waking up in the morning is a prerequisite condition for hitting the road for job.)
- Antonymy: It is a relation that one word is the opposite concept of an another word. (For example the word "happy" is the antonym of the word "sad".)

This dictionary model may be used in many forms and has advantages as compared with the use of specific WordNet synset relations. For example, if we want to relate words to one another and try making analogies between those words, the use of dictionary would be very helpful. Through implementation of clustering, many meaningful clusters can be built, because the analogies between the properties of words (such as the words 'finger' and 'hand' have meronymy relation and those would be in same cluster) may connect them in a semantic relationship. But instead of clustering process, we followed a simple statistical method, which would yield successful results.

## 3.2. Dictionary Preprocessing

In WordNet, relations are held in specific synsets, that is they have a structure based on word to word or word to noun-phrase relationships. So no parsing or disambiguation is needed for WordNet since words are in their root forms. However in basic dictionaries, there are no specific data structures that separately hold different synset relations. For example the meaning text of the word "cat" is as follows:

*"A small carnivorous mammal (Felis catus or F. domesticus) domesticated since early times as a catcher of rats and mice and as a pet and existing in several distinctive breeds and varieties."*

In the above example, the noun "rats" must be handled as the word "rat", whereas "mice" should be normalized as "mouse". The words should go through tokenization, stemming and normalization processes and then be taken into account. English is a language that has no much complexity concerning stemming, but as for Turkish, it is an agglutinative language and the parsing and disambiguation processes would quite matter. We used the parser and disambiguator tools, that are BoMorP and BoDis, developed for Turkish by Hasim Sak, Boğaziçi University [11, 12]. BoMorP implements the parsing operation and analyses the inflectional and derivational morphemes of a word and proposes part-of-speech (POS) tags. BoDis disambiguates amongst those POS tags and returns the one having the highest score according to an averaged perceptron-based algorithm.

A concept may be concerning a general idea of an abstract or concrete object, so the most probable concepts are generally nouns. So in this study we assumed that nouns represent concepts more than do any other word categories, so we took only nouns as concepts in the documents into account.

## 3.3 Document Analysis using Dictionary

A dictionary item may have many meanings and we have to determine which meaning is used in the context of a document. For example the word "cat" has many meanings and if the document contains this word we have to extract which meaning of this word is used in the document we are handling. For this, we looked up in the dictionary meaning text nouns as well as the nouns in the context of the word in the document and took the one yielding the highest similarity measure that has most common words. This formula is as follows:

$$\arg\max_m Similarity(m, c_w) = CommonCount(m, c_w)$$

In this formula, $m$ denotes a specific meaning of a word in a dictionary, whereas $C_w$ denotes the context of a word in a document. In this work, we used a context size of 30 words, such that 15 words that are on the left and 15 words that are on the right of the candidate word are taken into account. If amongst those words, many occurrences are encountered also in a specific meaning of the word to be disambiguated, that meaning would be selected as the true meaning. We also normalized the score (CommonCount) by dividing it by the number of nouns in the dictionary meaning text to get more accurate results.

## 3.4. Mapping Concepts

### 3.4.1. Simple Frequency Algorithm

A concept that can be extracted from a document has generally to do with frequency of the specific word encountered in that document. For example, if we encounter a document that abounds with the word "football", we may be inclined to think that the concept of this document would be concerning "sport". Taking frequency into account, we developed a statistical method, that extracts concepts favouring the words that are more frequent.

We first take all the nouns in the document(s) and label them as pre-concepts. Here we eliminate other types of words, such as adjectives and verbs. Then we start building a matrix. This matrix has rows representing the nouns encountered in the document and columns representing the nouns encountered in the meaning text sentences of those row words. But we also have to take into account that the rows on the words are also added as column items. For example the word 'football' may be very frequent in the document, so this word should be regarded as a probable concept as well.

The cells in the matrix are filled as follows: After we built the matrix, we fill the cells by 1 or 0 depending on whether the column word appears in the row word's meaning text or not. Then we implement frequency operation: We multiply all the cell values in the matrix by the corresponding row word frequency. For instance, if the word "football" is encountered 10 times in a document and its meaning text nouns in the dictionary are "sport", and "team", then those columns' ("sport", "team" and "football") values in the correspondent row "football" would be updated as 10, whereas the other columns would be updated as 0.

Then we summed up the column values in the matrix and took the column word that yields the highest summation as the probable concept. This is meaningful since the concept may or may not be present in the document, so the use of dictionary would be beneficial. An example showing the mapping of terms into concepts is shown in Figure 2.

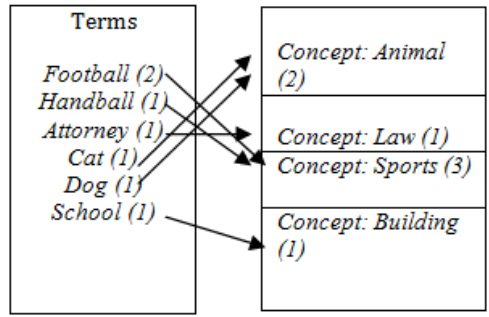

Figure 2. An example showing the mapping of words in a document into probable concepts.

In the above example, the column to the left is a representative of words encountered in the document, whereas the column to the right includes the representative nouns encountered in their meaning texts. Due to that the words "football" and "handball" are frequent in the above example and the word "sports" is present in their meaning texts, its score would be three and this word would be assigned as the probable concept of the document or corpus.

As mentioned above, we benefit from dictionary to extract concepts from documents, but instead of using just meaning text nouns for this concept extraction process, we also built a hierarchical data structure that contains 2, 3 and 4 levels. In accordance with this structure, the main word is atop the hierarchy, then the meaning text nouns of this word is in the lower level, whereas the respective meaning text nouns of these meaning text nouns are in the lower levels. An example of this data structure with 3-levels is depicted in Figure 3.

This hierarchical structure may have some specific features, for example each word in different levels may be assigned a different coefficient and we may take this coefficient factor into account when building up the matrix. If we construct 3-level hieararchies built through the dictionary, we may assign high values for the top levels and low values for lower levels. This is the case because the semantic relationship between the main word and the lower level nouns weakens while going down through the hierarchy structure. We multiplied the top-level words in the matrix by 1, the second-level words by 0.5 and the lowest-level words by 0.25. We used this geometric approach since the meaning text nouns' frequencies increase geometrically from one level to the below one. But we noticed that 3-level structure gives a very low success rate, so we preferred 2-level structure with no coefficients yielding high accuracy values. Also 4-level structure gave even worse results than did 3-level one, so using a 2-level structure was the best choice.

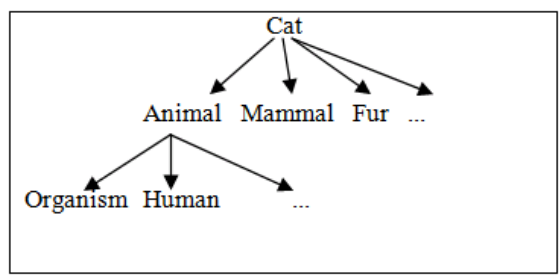

Figure 3. A hierarchical data structure with 3-levels of the word 'cat' in the dictionary.

We filled the matrix cells, as mentioned above, without taking frequency into account and saw that the results yielded were much less successful. That shows the importance of taking frequency into account.

We also have to take into account that some words are quite common in the dictionary, such as "situation", "thing", "person" and so on. Here we determined the top 1% most frequent words in the meaning texts in the dictionary as stop-words and eliminated them. Generally tf-idf is used for elimination of words, but since we make use of the dictionary as a base, top words elimination is sufficient.

### 3.4.2. Context and Frequency Algorithm

Although we noticed the algorithm we developed stated above gave meaningful concepts, drawbacks can be clearly seen. For example, let's assume there is a document containing the noun "football" and there is no other noun and its meaning text in the dictionary is as follows:

*"A game played by two teams of 11 players each on a rectangular, 100-yard-long field with goal lines and goal posts at either end, the object being to gain possession of the ball and advance it in running or passing plays across the opponent's goal line or kick it through the air between the opponent's goal posts."*

According to the algorithm stated above (Section 3.4.1), we take the nouns in this meaning text into account and build up a matrix containing those nouns, including the word 'football'. Since the word "football" is seen 3 times, the column labeled "goal" has a value of 3 as well and at the end the probable concept may be the word "goal", as well as the other concepts may be "game", "team", "line" and other nouns in the meaning text. (This is the case since the matrix would be of size 1 x *CountNoun(MeaningTextOf(Football))*, indicating that there is only one noun, that is "football", in the document.) Having a concept of "goal" through this document would be a bit nonsense, hence we modified the algorithm in the following manner:

All the dictionary meaning text nouns would not be useful in determining the general idea of the main word, so some of those nouns have to be eliminated. In order to determine which meaning text noun is relevant in the use of the main word, we used a corpus-based context analysis. We had a few corpora and for each corpus, we used a 30-word window size context analysis, that is we looked up 15 words on the left of the test words and 15 words on the right of the test words. Hereby we eliminated the context words which are not nouns, because we think of concepts as only nouns. Then we assumed that if a context word is also present in the meaning text of the main word in dictionary, we take this context word into account. After scanning the whole corpus, whichever context word is seen most, given that context word is also seen in the meaning text of the main word, we add this word as a column word in the matrix corresponding to the row word. Then, similar to what we have done in (Section 3.4.1), we multiply the row elements values by the frequency of the row representative word and sum up the columns values. Whichever column value has the maximum value, we define that column representative word as the probable concept. In this case, we take mostly two words for each word in the document: The word itself and the word in the meaning text of this word that is most widely seen in the contexts in the corpus. We again, of course, firstly eliminated the stop words present in the TDK Dictionary.

This approach makes sense, since all meaning text nouns would not be useful in determining the general idea, that is concept, of a word. Also the corpus-based approach shows that the most relevant word in the meaning text of a test word is extracted through the context analysis. Selecting at most two words, that are the word itself and the most frequent word in the contexts of the word that is also present in the meaning text of the row word in the matrix rather than taking into account all nouns in the meaning text of a word increased the success rate for three of the corpora.

## 4. EVALUATION AND RESULTS

We tested our algorithms on four corpora in Turkish, which are as follows: Gazi University, Sport News, Forensic News and Forensic Court of Appeals Decisions. The corpora we handled are in different fields of topics as follows:

- **Gazi University Corpus:** This corpus includes 60 text files. There is no specific topic in this corpus, instead the files collected represent different fields of topics, such as tutorials on specific architectural or engineering subjects and many others.
- **Sport News Corpus:** This corpus includes 100 text files. Each file has a topic concerning sport news, especially Turkish football.
- **Forensic News Corpus:** This corpus includes 100 text files. These files are collected from news that are concerning forensic domain.
- **Forensic Court of Appeals Decisions Corpus:** This corpus has 108 text files. These files are concerning the forensic decisions and this corpus is similar to the corpus Forensic News. The difference is that this one is not collected from news documents database.

Initially we parsed and disambiguated the files in those corpora through the tools BoMorp and BoDis to extract the nouns needed for concept mining. Then we discarded the stop-words that abound in the dictionary.

### 4.1. Evaluation Metric

As an evaluation metric, we made use of the accuracy metric, which can be defined as follows:

$$accuracy = \frac{true\ positive + true\ negative}{true\ positive + false\ positive + true\ negative + false\ negative}$$

In order to evaluate the concepts extracted through the algorithm we developed, we also manually determined the concepts in those corpora. Then we compared the concepts with one another. We handled many ways to compare the concepts suggested by the algorithm with the ones extracted manually: For documents we made comparisons in 3, 5, 7, 10 and 15 window sizes. These windows include the top concepts, for example the one with size 5, includes the top 5 concepts. A comparison example table is shown below:

Table 1. An example showing the top 3 concepts in 2 documents.

| Documents | Algorithm | Manual |
|-----------|-----------|--------|
| Document 1 | Sport, Game, Match | Sport, Match, Politics |
| Document 2 | Court, Attorney, Judge | Attorney, Accused, Match |

Table 1 shows the top three concepts for two documents, extracted both manually and algorithmically. In the first document, it can be seen that the success rate is 2 / (2 + 1) = 0.66, since there are two words in common, which are "sport" and "match" that are found both in concept clusters extracted manually and algorithmically. However the word "game" is not in the top three concept cluster yielded manually, so it decreases the success rate. In Document 2, the success rate is 0.33, since only the word "attorney" is common amongst the three top concepts.

We also made comparison in a manner which compares the top 3, 5, 7, 10 and 15 concepts found through the algorithm with all the concepts found manually and evaluated the success ratio. It generally gave the highest success rates.

## 4.2. Results

Table 2 summarizes the accuracy results for different corpora. Window sizes 3, 5, 7 and 10 are tested for comparison. For algorithm 1 (Simple Frequency Algorithm) the most successful results are achieved through method, which uses 2-level structure taking frequency into account. All values shown in Table 2 for algorithm 1 are that achieved by 2-level structure taking frequency factor into account. For this algorithm, the sub-methods taking 3-level structure into account or eliminating frequency factor gave unsuccessful results. It is seen, on average, comparison with window size of three gives the highest success rate. The second algorithm gives, on average, better evaluation results. It can clearly be seen that results for different corpora vary a lot which shows that concept mining would be biased for documents concerning specific topics. The first algorithm gives, on average, a success rate of 63.62%, whereas the second one gives a success rate of 75.51%.

Table 2. Performance results for different corpora in terms of accuracy percentage.

| Corpora | Comparison Window Size | Simple Frequency Algorithm | Frequency and Context Algorithm |
|---|---|---|---|
| Gazi | k = 3 | 58.40 | 69.30 |
| | k = 5 | 55.70 | 67.00 |
| | k = 7 | 54.70 | 64.30 |
| | k = 10 | 53.90 | 62.60 |
| Sport News | k = 3 | 57.40 | 55.40 |
| | k = 5 | 56.90 | 54.20 |
| | k = 7 | 56.00 | 53.40 |
| | k = 10 | 55.30 | 52.68 |
| Forensic News | k = 3 | 58.74 | 77.53 |
| | k = 5 | 56.81 | 71.72 |
| | k = 7 | 55.52 | 68.53 |
| | k = 10 | 55.18 | 67.79 |
| Forensic Decisions | k = 3 | 76.81 | 95.74 |
| | k = 5 | 67.45 | 91.54 |
| | k = 7 | 63.21 | 84.86 |
| | k = 10 | 59.44 | 78.95 |

Table 3 shows, as an example, the results for Forensic Decisions corpus comparing all methods. First Algorithm (2-levels, 1-0) is the method that fills the matrix with values 1 or 0 using 2-level structure. If a column word is present the cell value in matrix is 1, if it is not present, cell value is 0. Frequency factor is overlooked in this sub-method. First Algorithm (2-levels, frequency) takes frequency into account. First Algorithm (3-levels, coefficients) takes frequency into account using 3-level structure. This algorithm assigns different scores for different hierarchical levels of a word. The second algorithm is the algorithm explained in Section 3.4.2.

# 5. CONCLUSION AND FUTURE WORK

In this work, we developed two novel algorithms which make use of the dictionary. So far, generally WordNet has been preferred for its synset relations, especially hypernymy, in concept mining, because the general idea of a word is generally related to its hypernym word. But only taking a solitary synset relation into account may be insufficient, also other relationships may give us an idea concerning a word, that is why we made use of a general language dictionary.

Table 3. Performance results of four methods for Forensic Decisions corpus with different comparison windows sizes.

| Algorithms | Comparison Window Size | | | |
|---|---|---|---|---|
| | k = 3 | k = 5 | k = 7 | k = 10 |
| First Algorithm (2-levels, 1-0) | 76.81 | 67.45 | 63.21 | 59.44 |
| First Algorithm (2-levels, frequency) | 68.84 | 65.43 | 66.72 | 63.92 |
| First Algorithm (3-levels, coefficients) | 64.81 | 60.73 | 57.87 | 55.72 |
| Second Algorithm | 95.74 | 84.77 | 78.58 | 73.11 |

Besides the use of a dictionary, we took the frequencies into account, because the more frequent a word in the document, the higher contribution of this word to the general idea and concept of documents amongst many. We also developed context-based algorithm which eliminates some of the meaning text nouns in dictionary from probable candidate concepts set, because some of those words would not be convenient in determining the general idea, that is concept, of a word. This approach increased our success rate.

As a future work, we would improve the second algorithm, where we can take not only the word and the noun that is most frequently found in its contexts in all corpus and present in its meaning text in dictionary into account, but also the other nouns present in both meaning text and contexts. We can give them coefficients in accordance with their frequencies instead of eliminating them. Apart from the direction of this paper's algorithm, for concept mining we may also implement clustering through all dictionary words, in accordance with the analogies between their meaning texts. This would approach the whole dictionary word-set as a training data.

## ACKNOWLEDGEMENT

This work was supported by the Boğaziçi University Research Fund under the grant number 5187, and the Scientific and Technological Research Council of Turkey (TÜBİTAK) under the grant number 110E162. Cem Rıfkı Aydın is supported by TÜBİTAK BİDEB 2210. We thank Hasim Sak for providing us the tools for pre-processing and morphological analyses.

## Authors

Cem Rıfkı Aydın received his B.Sc. degree from Bahçeşehir University, Istanbul, Turkey, in Department of Computer Engineering. He is currently an M.Sc. student in Department of Computer Engineering at Boğaziçi University, and awarded with TÜBİTAK BIDEB scholarship throughout his M.Sc. studies. His areas of interest include natural language processing, artificial intelligence applications, pattern recognition, game programming, information retrieval and genetic algorithms.

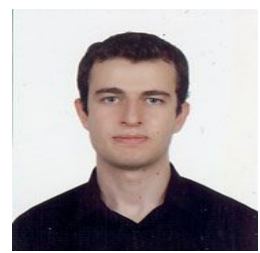

Ali Erkan received his B.Sc. and M.Sc. degrees from Department of Industrial Engineering, Bilkent University, Ankara, Turkey, and he received M.Sc. degree in Software Engineering from Department of Computer Engineering, Boğaziçi University, Istanbul, Turkey, in 2010. He is currently studying for Ph.D. degree at Department of Computer Engineering, Boğaziçi University, Istanbul, Turkey. His research interests include natural language processing, machine learning, pattern recognition, bioinformatics and statistics.

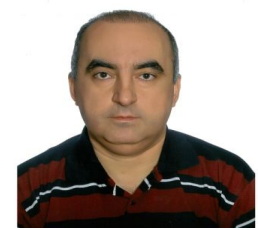

Tunga Güngör received his Ph.D. degree from Department of Computer Engineering, Boğaziçi University, Istanbul, Turkey, in 1995. He is an associate professor in the department of Computer Engineering, Boğaziçi University. His research interests include natural language processing, machine translation, machine learning, pattern recognition, and automated theorem proving. He published about 60 scientific articles, and participated in several research projects and conference organizations.

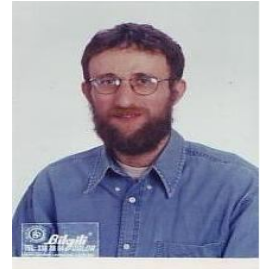

Hidayet Takçı is an academician at Cumhuriyet University, Sivas, Turkey. He studies on some fields such as data mining, text mining, machine learning and security. Hitherto he has lectured some courses such as data mining and applications, text mining, neural nets, etc. In addition, he has many papers and projects in the field of text classification and text mining, information retrieval and web security.

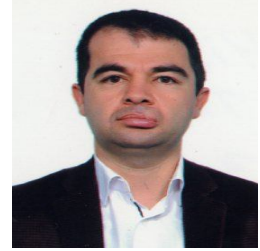